\RequirePackage{amsmath}
\documentclass[runningheads]{llncs}
\usepackage{amssymb}
\usepackage[T1]{fontenc}
\usepackage{times}
\usepackage{latexsym}
\usepackage[usenames,dvipsnames]{color}

\usepackage{multicol}
\usepackage{float}
\usepackage{graphicx}

\usepackage{array}
\usepackage{makecell}
\usepackage{enumitem}
\usepackage{algorithm2e}
\usepackage{mathtools}
\usepackage{wrapfig}
\usepackage{hyperref}

 \usepackage{CJKutf8}

\usepackage{todonotes}
\usepackage{xcolor}
\definecolor{brightpink}{rgb}{1.0, 0.0, 0.5}

\begin{document}

\RestyleAlgo{ruled}
\title{Kastor: Fine-tuned Small Language Models for Shape-based Active Relation Extraction}
%
\titlerunning{Kastor: Fine-tuned SLMs for Shape-based Active Relation Extraction}
%
\author{
  Célian Ringwald\inst{ 1}\orcidID{0000-0002-7302-9037},
     Fabien Gandon\inst{1, 2}\orcidID{0000-0003-0543-1232},
     Catherine Faron\inst{1}\orcidID{0000-0001-5959-5561},
     Franck Michel\inst{ 1}\orcidID{0000-0001-9064-0463},
    Hanna Abi Akl\inst{ 1, 2}\orcidID{0000-0001-9829-7401}
}
\authorrunning{Ringwald et al.}
%
\institute{Université Côte d’Azur, Inria, CNRS, I3S \and
 Data ScienceTech Institute
}

\maketitle              
\begin{abstract}
RDF pattern-based extraction is a compelling approach for fine-tuning small language models (SLMs) by focusing a relation extraction task on a specified SHACL shape. This technique enables the development of efficient models trained on limited text and RDF data. In this article, we introduce Kastor, a framework that advances this approach to meet the demands for completing and refining knowledge bases in specialized domains.
Kastor reformulates the traditional validation task, shifting from single SHACL shape validation to evaluating all possible combinations of properties derived from the shape. By selecting the optimal combination for each training example, the framework significantly enhances model generalization and performance. Additionally, Kastor employs an iterative learning process to refine noisy knowledge bases, enabling the creation of robust models capable of uncovering new, relevant facts.
\end{abstract}
%
%
%
\section{Introduction}

Relation extraction (RE), the task of retrieving relations from unstructured text, was drastically improved by language models and massive corpora aligning texts and facts from Knowledge Bases (KB) -- e.g. Wikipedia articles with corresponding Wikidata or DBpedia subgraphs. The use of generative seqToseq models is prevalent today to solve structured output generation, as they are flexible compared to encoder-only models, which require a decoding strategy. In this perspective, fine-tuned encoder-decoder models such as T5 or BART demonstrated good performances in the relation extraction domain~\cite{paolini2021structuredpredictiontranslationaugmented,dligach-etal-2022-exploring,rossiello2022knowglknowledgegenerationlinking}, and subsequent works underlined that small models could compete with larger models~\cite{zaratiana-etal-2024-gliner,Li2023EvaluatingCI,DBLP:conf/esws/LehmannMMORSV24}. However, these RE models fine-tuned on large-scale datasets do not guarantee good results on rare relations and entities. In addition, the catastrophic forgetting effect~\cite{li-etal-2022-overcoming} makes difficult a second-step adaptation of these models to a more specific domain. On the other hand, distant supervision approaches~\cite{10.1007/978-3-319-13704-9_3,10.1145/3241741} used to align massive text corpora and facts from databases have two significant drawbacks: (1) they may incorporate noise at the model learning stage by coupling a graph describing facts not present in the text and vice versa, and (2) they also affect how the resulting models are evaluated, from a knowledge completion point of view, by labelling as false positives results containing relevant information.

\noindent \textbf{Kastor (Knowledge Active Shape-based extracTOR)} answers these issues by proposing the refinement of specialized and frugal small language models (SLMs) focused on specific RDF patterns. The resulting model can produce structured data that could populate a database directly. Building on previous work~\cite{10.1007/978-3-031-78952-6_8}, we propose here to characterize the combination of properties  $\mathbb{P}(s^*)$ derived from a given SHACL shape $s^*$. We reused the shape introduced in~\cite{10.1007/978-3-031-78952-6_8} targeting the \texttt{dbo:Person} class, as it represents $1/6^th$ of the DBpedia content, by targeting the 7 datatype properties that are most likely to be found in the abstracts. This shows that Kastor already scales.
Our framework consists of coupling an initial incomplete and noisy knowledge base $\mathcal{K}$ with language models, and it works in two stages. It first consolidates $\mathcal{K}$ by ensuring strong alignment of the dual base, which will then be sampled to obtain smaller, higher-quality training and evaluation datasets used for training an initial SLM. Kastor's second step integrates a light, active learning process involving a human annotator. This annotator is used to analyze and correct the outputs produced by this first model to produce a gold model, which could extract relevant RDF graphs to complete $\mathcal{K}$. To summarize, this paper addresses the two following research questions: 
\begin{itemize}
    \item \textbf{RQ1.} To what extent does a task relying on example-specific achievable patterns improve the performance of relation extraction model?
    \item \textbf{RQ2.} Does an active learning process enhance relation extraction models?
\end{itemize}

\section{Related Work}\label{sec:related_work}

\textbf{SLMs are competitive}. The usage of LLMs is called into question today~\cite{grangier2024needsmallspecializedlanguage,wang2024comprehensivesurveysmalllanguage,lu2024smalllanguagemodelssurvey}, as they are costly to train, slow in inference and hard to adapt to a specialized domain~\cite{10.5555/3618408.3619049}. Moreover, although LLMs are highlighted for their few-shots abilities, recent work~\cite{zaratiana-etal-2024-gliner} focused on NER challenged this belief. In the context of relation extraction, several works underline the good performances of fine-tuned SLMs over prompted LLMs ~\cite{10.1007/978-3-031-60626-7_11,DBLP:conf/text2kg/GallardoCCHB24,gahnem2024}. Finally, LLMs only marginally outperform SLMs with the help of great engineering work~\cite{Efeoglu2024RetrievalAugmentedGR,zhang2023usinglargelanguagemodels,ijcai2024p0704,wadhwa-etal-2023-revisiting}.

\noindent\textbf{Human feedback is costly but necessary}. Because of the noise produced by distant supervision methods, the two classical datasets related to the relation extraction task, TACRED and DOCRED,  were revised and corrected several times~\cite{Stoica2021ReTACREDAS,alt-etal-2020-tacred,yao-etal-2019-docred,yao-etal-2021-codred}. Wikidata and DBpedia are, by construction, both concerned with coverage and quality issues~\cite{SHENOY2022100679,hofer2023constructionknowledgegraphsstate}, and the datasets recently proposed, such as TREX and REBEL do not spare these issues. For this reason, several works integrate partial human annotation to reduce noise in training datasets~\cite{huguet-cabot-etal-2023-red,ma-etal-2023-dreeam}. The recent concept of LLM-as-judge aims to replace human intervention with LLMs~\cite{zheng2023judgingllmasajudgemtbenchchatbot}. Nevertheless, this proposal is imperfect, and \cite{Li_2023} proposed combining both approaches (CoAnnotation). \cite{vandermeer2024annotatorcentricactivelearningsubjective,tsaneva_enhancing_nodate} also demonstrated the potential of this synergy. \\
\noindent\textbf{The problem of hallucination}. The literature on hallucination shows how language models fail to produce expected values. A first dichotomy is proposed in~\cite{Ji_2023} between the intrinsic hallucination (which contradicts the input) and the extrinsic hallucination (which cannot be verified by the input). \cite{huang2023surveyhallucinationlargelanguage} goes further in this categorisation by differentiating factuality hallucination, which relates to outputs that could not be supported by the text, and faithfulness hallucinations, which is related to the consistency of the retrieved answer, considering contextual and logical aspects. In the context of the NER task~\cite{Rogulsky2024TheEO} demonstrates that the noise of the training set directly impacts the hallucination rate of fine-tuned SLMs.

\noindent\textbf{Positioning:} The shape-based RE extraction framework we propose is a new task that is difficult to compare with the RE state-of-the-art. First, RE models based on the finetuning of PLM offer no control over the properties to extract, and they all generate triples following different linearizations~\cite{huguet-cabot-navigli-2021-rebel-relation,josifoski-etal-2022-genie,rossiello2022knowglknowledgegenerationlinking,paolini2021structuredpredictiontranslationaugmented} that would require additional data transformations that may create additional errors. Second, LLMs, despite their few shot capacities, demonstrate issues when it comes to producing structured output~\cite{geng-etal-2023-grammar,liu_are_2024}.

\noindent \textbf{Contributions:} (1) Kastor extends~\cite{10.1007/978-3-031-78952-6_8} by refining the task definition, and it demonstrates better performance in a wider variety of cases. (2) Kastor integrates this new task definition into a generalized framework, allowing the systematization of shape-based extractors over the chain, from the sample selection to the annotation and later to the PLM finetuning. (3) Kastor proposes a light-active process to build gold datasets and unbiased models, increasing the produced graph's relevance for a KB completion scenario. We also propose a characterization of the errors that (a) allows us to check the generated triples  and (b) could be extended in other settings. The produced material is made open and reusable\footnote{\url{https://github.com/datalogism/Kastor}}: both 
the resulting models\footnote{\url{https://zenodo.org/records/14498940}} and the produced datasets\footnote{\url{https://zenodo.org/records/14382674}
}.

\section{The Kastor framework}
\label{sec:kastor}

\subsection{The original task: an extraction focused on a maximal target shape $s^*$}\label{sec:rdf_extraction_old}
We start by formally defining the relation extraction task introduced in~\cite{10.1007/978-3-031-78952-6_8}. It relies on a training set built from the dual base $\mathcal{K}$ defined from the set $\mathcal{W}$ of Wikipedia abstracts associated with the set $\mathcal{G}$ of DBpedia graphs describing ($desc()$) the same resource $e$:
\begin{equation} \label{eq:K}
\begin{split}
\mathcal{K} \coloneqq \{(w,g) \in \mathcal{W}\times\mathcal{G}, \exists \, e \in IRI \text{ such that } desc_{\mathcal{W}}(e)=w \: \wedge \: desc_{\mathcal{G}}(e)=g\}
\end{split}
\end{equation}

To ensure the quality of the training set, we considered the subset of $\mathcal{K}$ where all the graphs are valid against a SHACL shape $s^*$. We call this shape maximal, as it matches the largest pattern to be extracted. We note $g \models s^*$ this validation, and $\mathcal{K}_{s^*}$ the corresponding subset:
\begin{equation} \label{eq:K-s-star}
\mathcal{K}_{s^*} \coloneqq \{(w,g) \in \mathcal{K} \, , \; g \models s^*\}
\end{equation}

Finally, to reduce the noise in $\mathcal{K}_{s^*}$, due to mismatch between DBpedia graphs and Wikipedia abstracts, we focused only on the couples $(w,g)$, where the abstract $w$ entails the paired graph $g$, i.e. the triples of $g$ that can actually be extracted from the paired abstract $w$. We note it $w \models g$ and we denote the dataset by $\mathcal{K}_{s^*}^{\mathcal{W} \models}$:

\begin{equation} \label{eq:K-s-star_W}
\begin{split}
\mathcal{K}_{s^*}^{\mathcal{W}\models} \coloneqq \{(w,g) \in \mathcal{K}_{s^*}  \, , \; w \models g\}
\end{split}
\end{equation}

$\mathcal{K}_{s^*}^{\mathcal{W}\models}$ is used to train a model expected to predict, from an abstract $w$, a graph $\hat{g}$ valid against $s^*$. We denote by $\mathcal{M}$ this original model, and baseline:
\begin{equation} \label{eq:M}
\begin{split}
\mathcal{M}:\begin{dcases}
\mathcal{W} \rightarrow \mathcal{G} \\
w \mapsto \hat{g} \, , \; \hat{g}  \models s^* \wedge w \models \hat{g}
\end{dcases} 
\end{split}
\end{equation}
%

\subsection{Example-specific patterns and Rule-based graph augmentation}
\label{sec:rdf_extraction_new}
In practice, many abstracts are short and miss some properties defined as mandatory in  $s*$. A model not trained to manage such cases could be encouraged to extract relations from an abstract with missing information, leading to hallucinations. So instead of considering a single graph pattern entailed by the SHACL shape $s^*$ for the whole training set, we propose to consider the example-specific ``achievable'' graph patterns $\pi \in \Pi$ for each pair $(w,g) \in \mathcal{K} $.
We denote by $\mathcal{P}(g)$ the set of properties occurring in a graph $g$: 
\begin{equation} \label{eq:pattern-graph}
\begin{split}
\mathcal{P}: \left| 
\begin{array}{l}
\mathcal{G} \longrightarrow  \Pi =2^I\\
g \longmapsto \pi = \{p_i \, , \; \exists \,  (x,p_i,o) \in g \}
\end{array}
\right. 
\end{split}
\end{equation}
By extension, we denote by $\mathcal{P}(s)$ the set of properties 
that a shape $s$ constrains a graph with.  
We denote by $\mathbb{P}(g)$ the powerset of $\mathcal{P}(g)$, respectively by $\mathbb{P}(s)$ the powerset of $\mathcal{P}(s)$, that could be deduced from a given $g$, respectively with $s^*$. 
We call \textit{example-specific patterns} the elements of $\mathbb{P}(g)$ and $\mathbb{P}(s)$.

To compare a graph $g$ with an example-specific pattern $\pi$, we note $g \rightarrow \pi$ the fact that all the properties in $g$ are found in $\pi$; and we note $g \nrightarrow \pi$ its negation, i.e. the fact that at least one property of $\pi$ is not found in $g$. We note $g \leftrightarrow \pi$  the fact that all the properties in $g$ are found in $\pi$ and vice versa; and we note $g \nleftrightarrow \pi$ its negation. 
By extension, when it comes to compare $g$ with a shape $s$, we denote by $g \rightarrow s$ the fact that all the properties in a graph pattern expressed in a shape $s$ 
are found in $g$; and by $g \leftrightarrow s$ the fact that all the properties in $g$ are found in the graph pattern expressed in $s$ and vice versa.

We can now relax the constraint in the original task that all the graphs in the training set must be valid against a single common shape $s^*$ (see eq.~\ref{eq:K-s-star}), in order to define a new training set considering all the example-specific patterns derived from $s^*$:
\begin{equation} \label{eq:K-P-s-star}
\begin{split}
\mathcal{K}_{\mathbb{P}(s^*)} \coloneqq \{(w,g) \in \mathcal{K} ; g \rightarrow s_i ; s_i \in \mathbb{P}(s^*)\}
\end{split}
\end{equation}

Additionally, we propose to complement $\mathcal{K}_{\mathbb{P}(s^*)} $ by materializing the closure of a set of inference rules $\mathcal{R}$ applied to $\mathcal{G}$. 
The rationale is as follows: some basic reasoning tasks can easily be deduced from the data by simple inference rules (e.g. deducing a year from a date). Therefore, instead of expecting the language models to learn these basic rules, we apply them in a declarative manner, thus homogenizing the graph and making the learning process easier.
The resulting set is denoted $\mathcal{K}^{\mathcal{R}\models}_{\mathbb{P}(s^*)}$:
\begin{equation} \label{eq:K-R-entails}
\begin{split}
\mathcal{K}^{\mathcal{R}\models}_{\mathbb{P}(s^*)} \coloneqq \{(w,g')\, , \; (w,g) \in \mathcal{K}_{\mathbb{P}(s^*)} \,  \text{, where } g' \text{ is the result of applying } \mathcal{R} \text{ to } g\}
\end{split}
\end{equation}

We ensure the entailment of the graphs by their paired abstracts ($w \models g$)  as done in the original design:
\begin{equation} \label{eq:K-s-star_RW}
\begin{split}
\mathcal{K}^{\mathcal{WR}\models}_{\mathbb{P}(s^*)}  \coloneqq \{(w,g) \in \mathcal{K}^{\mathcal{R}\models}_{\mathbb{P}(s^*)}  \, , \; w \models g\}
\end{split}
\end{equation}

The resulting set $\mathcal{K}^{\mathcal{WR}\models}_{\mathbb{P}(s^*)}$ is  sampled to finetune a new generation of model $\mathcal{M'}$:

\begin{equation} \label{eq:M'}
\begin{split}
\mathcal{M'}:\begin{dcases}
\mathcal{W} \rightarrow \mathcal{G} \\
w \mapsto \hat{g} \quad ; \quad \hat{g}  \leftrightarrow \mathcal{P}(g) \wedge (w,g) \in \mathcal{K}^{\mathcal{WR}\models}_{\mathbb{P}(s^*)} 
\end{dcases} 
\end{split}
\end{equation}
To train the models $\mathcal{M}$ presented in eq.~\ref{eq:M}, and $\mathcal{M'}$ presented eq.~\ref{eq:M'}, we considered random samples $RD$ which are split into $RD_{train}$, $RD_{eval}$ and $RD_{test}$.
We note $\mathcal{\mathcal{M'}}_{RD}$, the model $\mathcal{M'}$ trained on dataset $RD_{train}$, evaluated on $RD_{eval}$ and tested on $RD_{test}$.\\

Finally, to characterize the set of distinct example-specific patterns that are actually present in a given dataset $D$, and compare the variety of patterns in each dataset, we define:
\begin{equation} \label{eq:pattern-set-k-s}
\begin{split}\mathbb{P}_{D}(s) \coloneqq \{\pi \in \mathbb{P}(s); \exists (w,g) \in D ; g \leftrightarrow \pi\}
\end{split}
\end{equation}
For instance $\mathbb{P}_{RD^1}(s^*)$ will represent the set of example-specific patterns that can be built from $s^*$ and actually found in $RD^1$.

\subsection{ Knowledge distillation: from $\mathcal{K}$ to $\mathcal{K}^{\mathcal{WR}\models}_{\mathbb{P}(s^*)}$}\label{sec:experiment_pipeline}
\begin{figure}[h]
 \centering
 \vspace{-20pt}
 \includegraphics[width=0.6\linewidth]{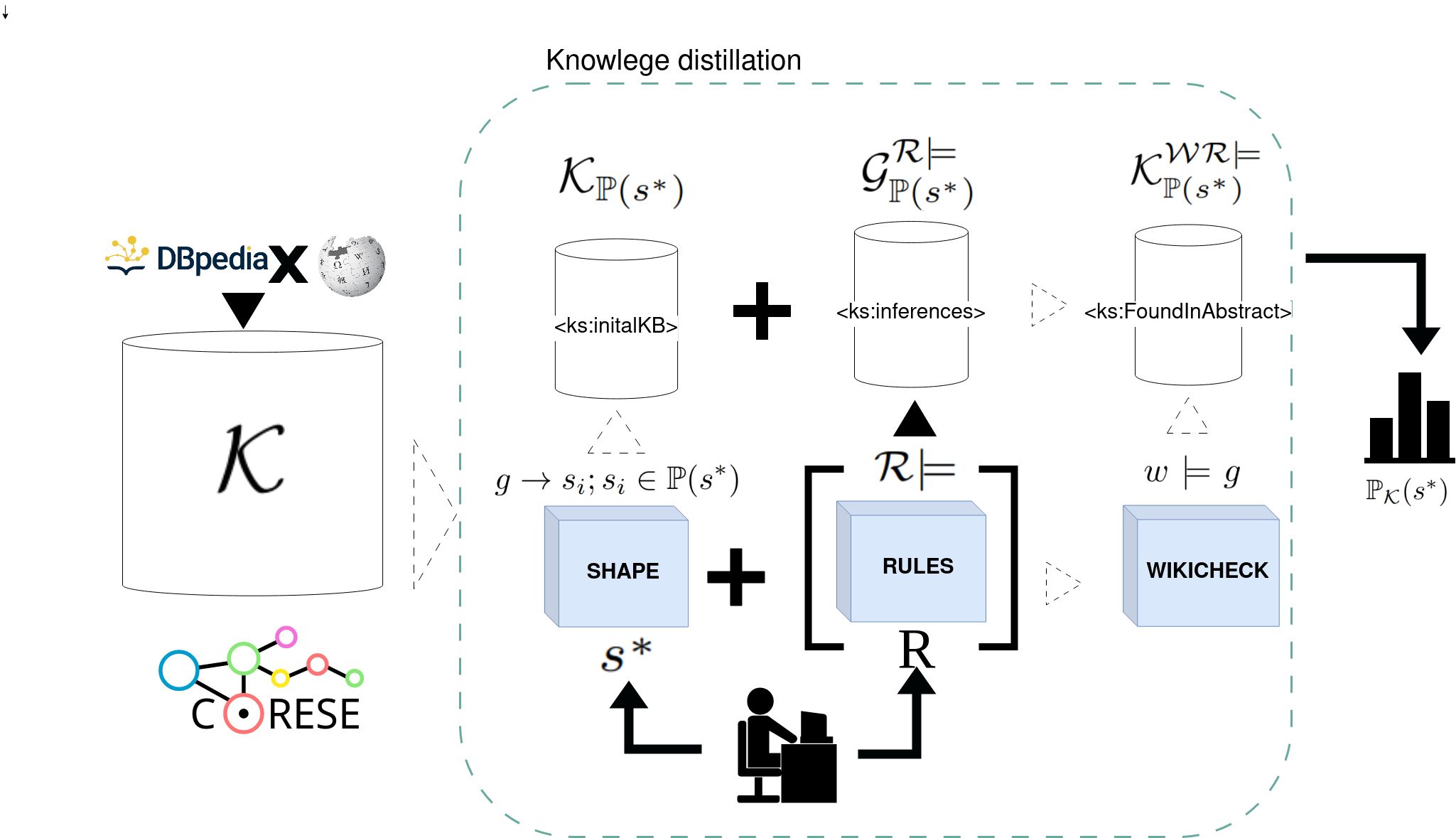}
 \caption{Knowledge distillation: from the initial base to a refined version based on shape $s^*$}
 
 \label{fig:kastor}
 \vspace{-10pt}
 \end{figure}

We consider the \textbf{dual base $\mathcal{K}$} consisting in the 2022.09 DBpedia datadump\footnote{\url{https://databus.dbpedia.org/cringwald/collections/kstor}}
that gathers 6.109.994 Wikipedia abstracts and their DBpedia graphs. 
We consider the \textbf{shape $s^*$} described in~\cite{10.1007/978-3-031-78952-6_8}\footnote{\url{https://github.com/datalogism/12ShadesOfRDFSyntax\#shacl-shape-used}} which targets instances of class \texttt{dbo:Person} and expresses that they must have one label, one birth year or date, and possibly 4 other optional properties: \\
$\mathcal{P}(s^*)=$ \begin{small}\{$\texttt{rdfs:label}, \texttt{dbo:alias},\texttt{dbo:birthName}, \texttt{dbo:birthDate}, \\\texttt{dbo:deathDate}, \texttt{dbo:birthYear},  \texttt{dbo:deathYear}$\}\end{small}\\ 
and from it we can compute $|\mathcal{P}(s^*)|=7$ and  $\left|\mathbb{P}(s^*)\right|=128$. 
There are 1.833.493 Person graphs in $\mathcal{G}$ valid against at least one of the 127 non-empty graph patterns in $\mathbb{P}(s^*)$. They are gathered in \textbf{the subbase $\mathcal{K}_{\mathbb{P}(s^*)}$} stored into a named graph \texttt{<ks:initalKB>}.
We observed in $\mathcal{G}_{\mathbb{P}(s^*)}$ a huge number of birth/death dates compared to the number of birth/death years. To solve this gap, we defined the following set of inference rules:
\begin{equation} \label{eq14}
\footnotesize
\mathcal{R}: \begin{dcases}
       \texttt{dbo:deathDate}  \models \texttt{dbo:deathYear} \\
       \texttt{dbo:birthDate}  \models \texttt{dbo:birthYear}    
\end{dcases}
\end{equation}
These rules, encoded as SPARQL Update queries, are applied to \texttt{<ks:initalKB>}, producing an \textbf{enriched base $\mathcal{K}^{\mathcal{R}\models}_{\mathbb{P}(s^*)}$}. 
We note $\mathcal{G}^{\mathcal{R}\models}_{\mathbb{P}(s^*)}$ the 900.000+ \textit{new} triples produced, and we store them in the named graph \texttt{<ks:inferences>}. 

We filter $\mathcal{K}^{\mathcal{R}\models}_{\mathbb{P}(s^*)}$ with the \textit{wikicheck} module in charge to verify eq.~\ref{eq:K-s-star_RW}. The \textbf{resulting base $\mathcal{K}^{\mathcal{WR}\models}_{\mathbb{P}(s^*)}$} is stored in the named graph \texttt{<ks:foundInAbstract>}. 
Statistics on the different graphs are given in Table~\ref{tab:data_consolidation}.
$\mathcal{K}^{\mathcal{WR}\models}_{\mathbb{P}(s^*)}$ filters 40\% of the entities described in $\mathcal{K}_{\mathbb{P}(s^*)}$ that have no property value in their graph that can be found in their abstract. This shows the importance of such filtering to avoid training models on facts unsupported by the  abstract, which could encourage hallucinations.
Note that some properties are often found in the abstracts, such as \texttt{dbo:birthYear}, or \texttt{dbo:deathDate}, whereas \texttt{dbo:alias} is often missing.

\begin{table}[h!]
\centering
\setlength{\tabcolsep}{0.7em}
\begin{tabular}{l|rrr|r}
predicate & \makecell{ $\mathcal{K}_{\mathbb{P}(s^*)}$  \\ \tiny{<ks:initialKB>} } &  \makecell{ $\mathcal{K}^{\mathcal{R}\models}_{\mathbb{P}(s^*)}$ \\ \tiny{<ks:initalKB> $\cup$ <ks:inferences>}   }& \makecell{ $\mathcal{K}^{\mathcal{WR}\models}_{\mathbb{P}(s^*)}$\\ \tiny{<ks:foundInAbstract>}    }  & Part Found \\
\hline
\texttt{dbo:birthName} & 120 102 & 120 102 & 84 377 & 70\%\\
\texttt{dbo:birthYear} & 186 575 & \textbf{960 977} & \textbf{888 745} & 92\%\\
\texttt{dbo:deathDate} & 250 970 &  250 970 & 213 424 & \textit{85\%}\\
\texttt{dbo:alias} & \underline{30 648} &  \underline{30 648} & \underline{10 571} & \underline{35\%}\\
\texttt{dbo:deathYear} & 97 817 &   335 571 & 315 671 & \textbf{94\%}\\
\texttt{dbo:label} & \textbf{987 389} &  \textbf{987 389} & \textit{596 286} & 60\%\\
\texttt{dbo:birthDate} & \textit{705 952} &  \textit{705 952} & \textit{605 239} & \textit{85\%}  \\
\hline
Nb. entities  &  1 833 493 & 1 833 493 & 1 093 886 &   60\%
\end{tabular}
\\
  \caption{Number of triples in each consolidated named graph }
  \label{tab:data_consolidation}
\vspace{-20pt}
\end{table}

\begin{figure}[h!]
\vspace{-10pt}
\begin{minipage}[t]{.6\linewidth}
\vspace{0pt}
\centering
\includegraphics[width=1\columnwidth,height=0.75\columnwidth]{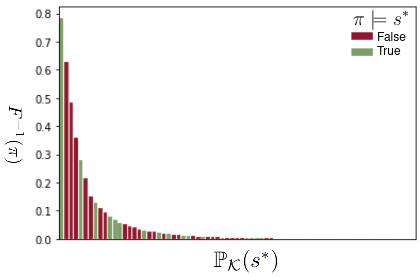}
\caption{$\mathbb{P}_{\mathcal{K}}(s^*)$ Inverse cumulative frequency distribution}\label{fig:longtail}
\end{minipage}
\begin{minipage}[t]{.4\linewidth}
\vspace{0pt}
\centering
\begin{tabular}{l|lll|llll|l|r}
idx & \rotatebox[origin=c]{90}{label} & \rotatebox[origin=c]{90}{birthDate} & \rotatebox[origin=c]{90}{birthYear} & \rotatebox[origin=c]{90}{deathYear} & \rotatebox[origin=c]{90}{alias} & \rotatebox[origin=c]{90}{birthName} & \rotatebox[origin=c]{90}{deathDate} & \rotatebox[origin=c]{90}{$s^*$ valid} &   freq\\ 
\hline
1 & \checkmark & \checkmark & \checkmark &  &  &  &  & \checkmark & 21.6\%\\
2 & \checkmark &  &  &  &  &  &  &  & 15.5\%\\
3 &  &  & \checkmark &  &  &  &  &  & 14.4\%\\
4 &  & \checkmark & \checkmark &  &  &  &  &  & 12.4\%\\
5 & \checkmark & \checkmark & \checkmark & \checkmark &  &  & \checkmark & \checkmark & 8.1\%\\
6 &  & \checkmark & \checkmark & \checkmark &  &  & \checkmark &  & 6.6\%\\
7 &  &  & \checkmark & \checkmark &  &  &  &  & 6.1\%\\
8 & \checkmark &  & \checkmark &  &  &  &  & \checkmark & 2.4\%\\
9 &  & \checkmark & \checkmark &  &  & \checkmark &  &  & 1.7\%\\
10 &  & \checkmark & \checkmark & \checkmark &  & \checkmark & \checkmark &  & 1.6\%\\
... & ... & ... & ... & ... & ... & ... & ... & ... & ...\\
\end{tabular}
\caption{10 most frequent $\pi \in \mathbb{P}_{\mathcal{K}}(s^*)$}\label{fig:longtail_data}
\end{minipage}
\vspace{-20pt}
\end{figure}
Reusing the notation from eq.~\ref{eq:pattern-set-k-s} applied to $\mathcal{K}^{WR\models}_{\mathbb{P}(s^*)}$, we get \textbf{an example-specific pattern set $\mathbb{P}_{\mathcal{K}}(s^*)$} containing $|\mathbb{P}_{\mathcal{K}}(s^*)|=70$ patterns, which is nearly half of the 127 possible patterns in $\mathbb{P}(s^*)$. The distribution of these patterns, shown in Fig.~\ref{fig:longtail}, is typical of a long-tail distribution. All these patterns can be classified as compliant or not with the maximal SHACL shape $s^*$ (cf. colours of the bars on Fig.~\ref{fig:longtail} and column ``$s^*$ valid'' in Fig.~\ref{fig:longtail_data}). 
There are only 47 patterns in $\mathbb{P}_{\mathcal{K}}(s^*)$ validating $s^*$ that can be found in the base. The original design of~\cite{10.1007/978-3-031-78952-6_8}, formalised in eq.~\ref{eq:M}, was for this reason focused on a subset of $\mathbb{P}_{\mathcal{K}}(s^*)$ (see eq.~\ref{eq:pattern-set-k-s}).

\subsection{An iterative SLM learning process with human in the loop}\label{sec:active_learning}

\subsubsection{Learning from a small dataset.}
In~\cite{10.1007/978-3-031-78952-6_8} we followed a 5-fold cross-validation training process implying 4 000 training examples, with  250 disjoint examples used for the evaluation and 1 000 test examples. We repeated the experiment to find a better cost-performance balance by testing different fold numbers and sample sizes. 
We empirically established that a 10-fold cross-validation based on a sample size of 1 000 rotating examples is sufficient to fine-tune an efficient model (cf. Table~\ref{tab:tab_config_ds}).
\begin{table}[h]
\centering
\vspace{-20pt}
\setlength{\tabcolsep}{0.8em}
\begin{tabular}{r|rrr|rl}
Nb. folds & Nb ex. train & Nb ex. test & Nb ex. eval & time & $F1^+$ \\
\hline
10 &  900 (90\%) & 100 (10\%) & 100 (10\%) & \textbf{2h} & 0.90 \\
5 & 800 (80\%) & 200 (20\%) & 200 (20\%) & 3h & 0.89 \\
5 & 4000 (80\%) & 1000 (20\%) & 250 (~5\%) & 5h & 0.91 \\
5 & 2000 (80\%) & 500 (20\%) & 500 (~5\%) & 6h & \underline{0.89} \\
10 & 2250 (90\%) & 250 (10\%) & 250 (10\%) & 7h & 0.93 \\
10 & 4500 (90\%) & 500 (10\%) & 500 (10\%) & 13h & \textbf{0.97} \\
5 & 4000 (80\%) & 1000 (20\%) & 1000 (10\%) & \underline{14h} & 0.91 \\
\end{tabular}
\caption{Impact of the number of folds and sample size on the model performances, sorted by computation time }\label{tab:tab_config_ds}
\vspace{-40pt}
\end{table}

\subsubsection{The sampler.} The sampler is the first piece of the Active Learning process; it selects random and independent subsets from $\mathcal{K}^{\mathcal{WR}\models}_{\mathbb{P}(s^*)}$. We used it to generate two datasets: $RD^0$ and $RD^1$ of 1200 examples (including 100 additional examples in case we remove some examples during the annotation phase). 
The sampler also generated an independent dataset of 600 examples, $RD^2$, used as a control. 
We also created the $RD^-$ dataset to serve as a baseline, reproducing the original model (eq.~\ref{eq:M}) against which to compare our model (eq.~\ref{eq:M'}). To obtain it we sampled $\mathcal{K}^{WR\models}_{\mathbb{P}(s^*)}$ with an additional filter selecting only pairs $(w,g)$ where $g$ is valid against $s^*$.

\subsubsection{SLM training details.}\label{sec:details_ft} All the trained models follow the design of~\cite{10.1007/978-3-031-78952-6_8} which is extended: For the pre-trained model, we chose BART-base (140M parameters). We linearized the graphs in the  TurtleLight online and factorized syntax, which demonstrated the best performances. The models were finetuned on a single Tesla V100-SXM2-32GB GPU using the same configuration: an inverse square root scheduler with an initial learning rate of 0.00005, 1000 steps of warmup, and configured with an early stop mode with patience of 5 steps. We also used the same prompt to finetune the models: ``\texttt{\$entity\_URI : \$Abstract}'' where \$Abstract is a Wikipedia abstract and \$entity\_URI the URI of the corresponding entity in DBpedia.
We extended the initial code with new metrics definitions, a validation of the TurtleLight syntax via an Extended Backus-Naur Form (EBNF) grammar, the validation of the produced triples against the expected $\mathbb{P}(s^*)$ patterns, the automation of the dataset construction, as well as the new testing process.
\subsubsection{Light active learning.} The process described in Appendix~\ref{annex:ActiveLearn} relates to a single loop annotation, creating a gold dataset and model. It starts by finetuning a first model $\mathcal{M'}_{RD^{0}}$ on $RD^0$. 
This model generates from the abstracts in $RD^1$ and $RD^2$  predicted graphs $\hat{g}$, which will be compared to the expected graphs $g$. We only keep the False Positives triples (FP) and False Negatives (FN) triples of $\hat{g}$ produced by testing $\mathcal{M'}_{RD^{0}}$ with $RD^1$ and $RD^2$. The annotator is a domain expert who has to evaluate the FP/FN triples collected regarding the Wikipedia abstract given in input to the model. A triple is considered erroneous if the datatype property value cannot be found in the text or if it is not strictly equal to the expected value.

Each dataset is then enriched with all the FP+ triples (the correct FP) and corrected by deleting the FN- triples (the erroneous FN) to produce two gold standard datasets: $RD^{1+}$ and $RD^{2+}$.

Each triple of $RD^{1+}$ and $RD^{2+}$ is evaluated with the NLI and the Triplet Critic models proposed in ~\cite{huguet-cabot-navigli-2021-rebel-relation} and ~\cite{huguet-cabot-etal-2023-red} and the scores are stored in dedicated named graphs (\texttt{<ks:sample1+>} and \texttt{<ks:sample2+>}) following the classical RDF design-pattern:\\
\begin{footnotesize} \texttt{ ?s ?p \_:n. \_:n rdf:value ?o; ks:triplet\_critic ?tc; ks:xnli ?nli.}\end{footnotesize} 

$RD^{1+}$ is finally used to produce a gold model $\mathcal{M'}_{RD^{1+}}$ and $RD^{2+}$ will serve as a control to compare the performances of $\mathcal{M'}_{RD^{0}}$ and $\mathcal{M'}_{RD^{1+}}$.

\begin{figure}[h!]
 \centering
 \includegraphics[width=0.8\linewidth]{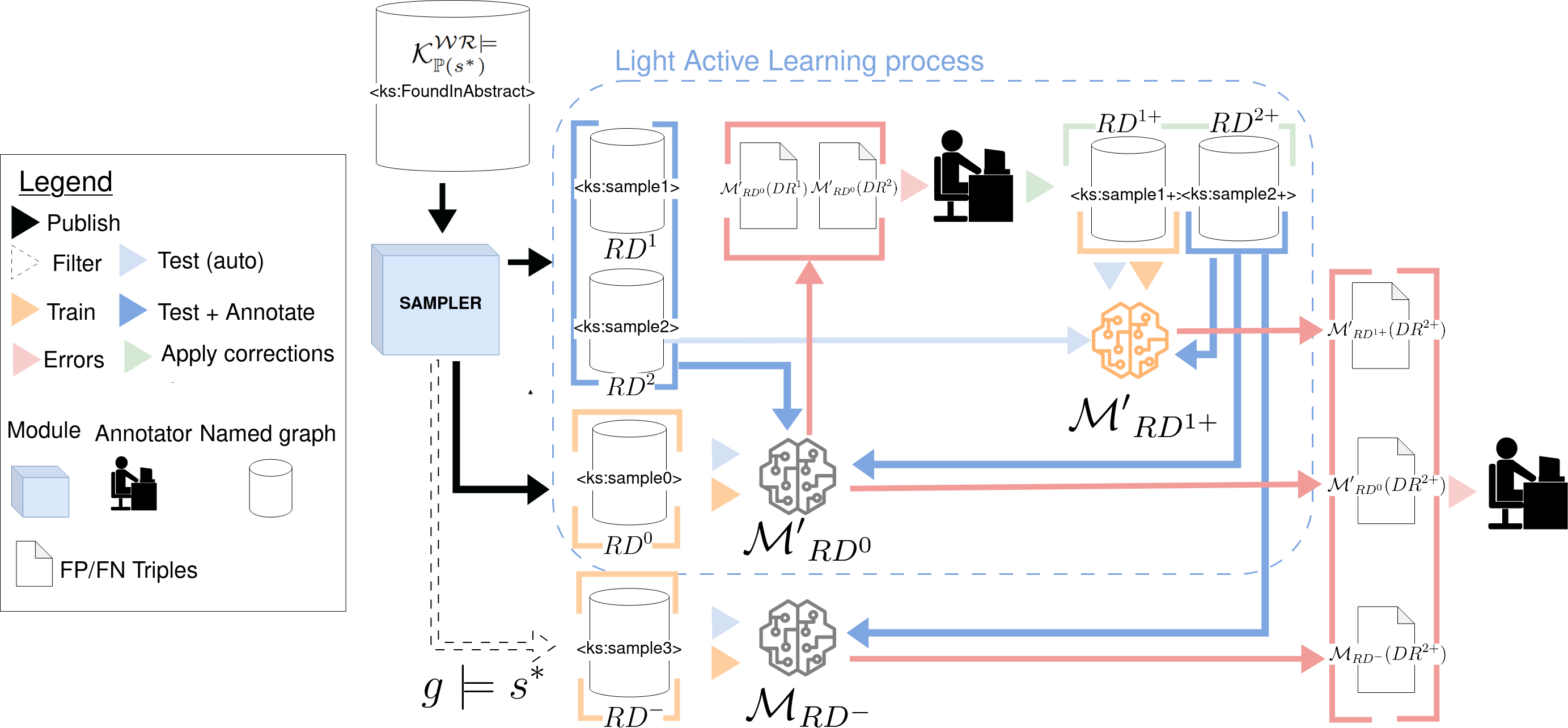}
 \caption{Overview the experimental set-up including the iterative SLM light active learning process}
 \label{fig:acl_kastor}
\vspace{-20pt}
 \end{figure}
\subsubsection{Cross-evaluation of the models.}
The complete experimental process followed in this work is presented and summarised in Fig.~\ref{fig:acl_kastor}. During the active learning process, we train two models: $\mathcal{M'}_{RD^{0}}$ and $\mathcal{M'}_{RD^{1+}}$ (cf. orange arrows in Fig.~\ref{fig:acl_kastor}). $RD^{1}$ and $RD^{2}$ are used as test datasets and then corrected after the annotation (cf. dark-blue arrows in Fig.~\ref{fig:acl_kastor}). We also tested $RD^{2}$ on $\mathcal{M'}_{RD^{1+}}$ to evaluate the gold model in a noisy context. Out of the active process, we train $\mathcal{M}_{RD^{-}}$.  After training, we test all these models with a dedicated subset of $D$: $D_{test}$ (cf. the light blue arrows in Fig 5). Finally, we also compare all the produced models with our gold dataset $RD^{2+}$ by annotating all the FP/FN they produced.

\section{Experimental results}\label{sec:results}
\subsection{Resulting Datasets}

We evaluate a dataset $D$ using several metrics. First, $\overline{|\mathcal{P}(g)|}$ is the average number of properties of the graphs $g$ in a dataset. $|\mathbb{P}_{D}(s^*)|$ is the number of distinct patterns descending from $s^*$ found in $D$. We also reused the NLI model\footnote{\url{https://huggingface.co/joeddav/xlm-roberta-large-xnli}}~\cite{huguet-cabot-navigli-2021-rebel-relation} and Triplet Critic\footnote{\url{https://huggingface.co/Babelscape/mdeberta-v3-base-triplet-critic-xnli}}~\cite{huguet-cabot-etal-2023-red} to estimate if an abstract $w$ contains the information needed to generate the triples of the associated $g$. These scores, noted $\overline{NLI(w,g)}$ and $\overline{TC(w,g)}$, are defined between 0 and 1. Finally we define $r_{s^*}$ as the rate of graphs following the maximal shape $s^*$:

\begin{equation} \label{eq:r_s_D}
r_{s^*}(D)=\frac{|D_{s^*}|}{|D|}
\end{equation}

\begin{table}[H]
\vspace{-20pt}
\centering
\setlength{\tabcolsep}{0.7em}
\begin{tabular}{l|llllll}
$D$ & $|D|$ & $\overline{|\mathcal{P}(g)|}$ & $|\mathbb{P}_{D}(s^*)|$ & $\overline{NLI(w,g)}$ & $\overline{TC(w,g)}$  & $r_{s^*}(D)$\\
 \hline
$RD^{-}$ & 1200 & \textbf{3.6} & \underline{16} & 0.42 & \textbf{0.91} & \textbf{1.00}\\
\hline
$RD^{0}$ & 1200 & \underline{2.9} & 35 & \underline{0.40} & 0.55 & 0.47\\
$RD^{1}$ & 1200 & \underline{2.8} & \textbf{39} & \underline{0.40} & \underline{0.54} & 0.47\\
$RD^{2}$ & 600 & \underline{2.8} & 30 & 0.42 & 0.55 & 0.49 \\
\hline
$RD^{1+}$ & 1200 & \textit{3.27} & \textit{38} & \textit{0.59} & \textit{0.75} & 0.59\\
$RD^{2+}$ & 599 & \textit{3.24} & 32 & \textbf{0.60} & \textit{0.75} & \textit{0.59}\\
\end{tabular}
\caption{Resulting datasets basic statistics}
  \label{tab:sample_init_analysis}
\vspace{-20pt}
\end{table}

Firstly, by considering $RD^{-}$  (the dataset that only contains graphs validating the maximal shape $s^*$), we can observe that  $|\mathbb{P}_{RD^{-}}(s^*)|=16$, which represents only a tiny variety of the patterns realized in the global KB (i.e. $|\mathbb{P}_{\mathcal{K}}(s^*)|=70$). However, $RD^{-}$ records a high $\overline{TC(w,g)}$ score compared to other samples. Additionally, the initial datasets ($RD^{0}$, $RD^{1}$ and $RD^{2}$) all share similar particularities:
around 2.9 properties per entity, they realize between 30 and 40 patterns, which is half of the total number of realized patterns of $\mathbb{P}_{\mathcal{K}}(s^*)$. Finally, the corrected datasets ($RD^{1+}$,$RD^{2+}$) count more properties per entity; compared to their previous versions, they record an increased rate of graphs valid against the maximal shape and higher $\overline{NLI(w,g)}$ and $\overline{TC(w,g)}$ scores.

\subsection{Models performances}

\begin{wraptable}{r}{6cm}
  \label{tab:train_res0}
\centering
\vspace{-20pt}
\setlength{\tabcolsep}{0.5em}
\begin{tabular}{l|lc} 
model & $\overline{CO_2} (g)$ & $\overline{time}$ (mins)  \\
\hline
$\mathcal{M}_{RD^{-}}$ & 0,018 & 7,65 \\
$\mathcal{M'}_{RD^{0}}$ & 0,019 & 8,11 \\
$\mathcal{M'}_{RD^{1+}}$ & 0,018 & 7,91 
\end{tabular}
\caption{Training cost of fine-tuned models}
\vspace{-20pt}
\end{wraptable}

We can first underline the frugality of the proposed method in terms of carbon cost and training time. Using \textit{carbontracker}\footnote{\url{http://carbontracker.info/}}, the training cost of these models is competitive: they require less than 10 minutes of training and the $CO_2$ footprint associated to their training is limited. 
To evaluate our models, we propose to define $\mathbb{G}_D$, the graphs expected given a dataset $D$, and $\widehat{\mathbb{G}_D}$, the set of the predictions obtained from an SLM. 
These sets allow distinguishing the subset $\widehat{\mathbb{G}}^{\text{parsed}}_{D}$ of all the predicted graphs that follow the EBNF Turtle Light grammar introduced in Section~\ref{sec:details_ft}, and the subset of $\widehat{\mathbb{G}}^{\text{URI+}}_{D}$ which contains the proper URIs to represent the subjects:
\begin{equation} 
\begin{split}
\widehat{\mathbb{G}}^{\text{parsed}}_{D} = \{ \hat{g} ; (w,g) \in D ; \hat{g} \text{ parseable} \}
 \\
\widehat{\mathbb{G}}^{\text{URI+}}_{D} = \{ \hat{g} ; (w,g) \in D ;  e \in IRI \text{ the entity described by graph } g ;\\ \hat{e} \in IRI \text{ the entity described by graph } \hat{g} ; e= \hat{e} \}
\end{split}
\end{equation}

We can now introduce the rate of predicted graphs that were correctly parsed: $r_{tll}$, as well as $r_{URI+}$, the rate of correct subject URIs produced:

\begin{equation} \label{eq:r_ttl_URI}
r_{tll}=\frac{|\hat{\mathbb{G}}^{parsed}_D|}{|\mathbb{G}_D|} \qquad r_{URI+}=\frac{|\hat{\mathbb{G}}^{URI+}_D|}{|\hat{\mathbb{G}}^{parsed}_D|}
\end{equation}

\noindent To measure
the ability of our models to produce graphs $\hat{g}$ containing exactly the expected property set $\mathcal{P}(g)$, we define the rate of strict property set equivalence: 
\begin{equation} \label{eq:G_left_right_arrow_D}
\widehat{\mathbb{G}}^{\leftrightarrow}_{D}  = \{ \hat{g} ; (w,g) \in D ; \hat{g} \leftrightarrow \mathcal{P}(g)\} \qquad \qquad r_{\hat{\mathbb{G}}^{\leftrightarrow}_D}=\frac{|\hat{\mathbb{G}}^{\leftrightarrow}_D|}{|\hat{\mathbb{G}}^{parsed}_D|} 
\end{equation}

\noindent The averaged $\overline{Loss(\hat{\mathbb{G}})}$ based on the cross-entropy helps us evaluate how confident our models are at predicting a given $\hat{g}$ from an abstract $w$. Moreover, we compute $F_1$ scores at the macro and micro levels ($F_1^+$, $F_1^-$) based on the strict~\cite{taille-etal-2020-lets} equality of the expected and generated graphs.\\
\noindent Finally, as one of our objectives is to consider and correct the False Positives and False Negatives generated by a model on a given dataset, we also define two final metrics:
\begin{equation} \label{eq:r_fp_r_fn}
r_{FP}\!=\!\frac{|FP|}{|FP|\!+\!|TN|\!+\!|TP|\!+\!|FN|} \qquad r_{FN}\!=\!\frac{|FN|}{|FP|\!+\!|TN|\!+\!|TP|\!+\!|FN|}
\end{equation}

\begin{table}[H]
\centering
\setlength{\tabcolsep}{0.5em}
\begin{tabular}{ll|ll|lll|ll|r}
model   & test set     & $r_{tll}$ & $r_{URI+}$ & $\overline{Loss(\hat{\mathbb{G}})}$ & $F_1^-$   & $F_1^+$     & $r_{FP}$ & $r_{FN}$ & $r_{\hat{\mathbb{G}}^{\leftrightarrow}_D}$\\
\hline
$\mathcal{M}_{RD^{-}}$ & $RD^-_{test}$  & 0.99 & 1.00 & \textbf{0.004} & \textbf{0.994} & \textbf{0.935}           & \underline{0.0036} & \underline{0.0021} & \textbf{0.97}      \\
$\mathcal{M}_{RD^{-}}$ & $RD^{2+}$ & 0.99  & 1.00 & \underline{0.27} & \underline{0.887} & \underline{0.697}           & \textbf{0.0717} &  \textbf{0.0355} & \underline{0.43}    \\
\hline
$\mathcal{M'}_{RD^{0}}$ & $RD^0_{test}$  & 0.99 & 1.00 & 0.01 & 0.978 & 0.916           & 0.0155 & \underline{0.0023} & 0.92      \\
$\mathcal{M'}_{RD^{0}}$& $RD^{2}$  & 0.99  & 1.00 & 0.07 & 0.927 & 0.733          & \textit{0.0448} & \textit{0.0162} & 0.73      \\
$\mathcal{M'}_{RD^{0}}$ & $RD^{2+}$ & 0.99  & 1.00 & 0.07 & 0.94 & 0.785           & 0.014 &  \textbf{0.0395} & 0.72    \\
\hline
$\mathcal{M'}_{RD^{1+}}$ & $RD^{1+}_{test}$ & 0.99 &  1.00 & \textbf{0.004} & \textbf{0.991} & 0.916      & 0.0054 & 0.0031 & \textit{0.95}   \\
$\mathcal{M'}_{RD^{1+}}$ & $RD^{2}$ & 0.99  & 1.00 & 0.12 & 0.907 & 0.724            & \textbf{0.0737} &  0.0076 & 0.62    \\
$\mathcal{M'}_{RD^{1+}}$ & $RD^{2+}$ & 0.99  & 1.00 & \textit{0.04} & \textit{0.958} & \textit{0.806}        & 0.026 & 0.0131 & 0.80      
\end{tabular}
\caption{Average performances of fine-tuned models over 10 folds on different test sets, bold values are the best recorded values, italic values represent the second best values, and the underlined values are the worst ones}
\label{tab:results_1}
\vspace{-20pt}
\end{table}
Table~\ref{tab:results_1} presents the results obtained in the same manner as the other tables in this paper. 
At first glance, we can notice that all the produced models generate close to syntactically perfect Turtle Light RDF graphs ($r_{ttl}$) and refer in every case to the correct focused entity URIs ($r_{URI+}$). The first model $\mathcal{M}_{RD^{-}}$ records high performances regarding the test set $RD^-_{test}$: a low $\overline{Loss(\hat{\mathbb{G}})}$, high $F1$ metrics and a significant part of produced graphs follow the expected patterns ($r_{\hat{\mathbb{G}}^{\leftrightarrow}_D}$).
Nevertheless,  $\mathcal{M}_{RD^{-}}$ completely failed on $RD^{2+}$,  which reveals an apparent lack of generalisation over graph patterns that do not strictly validate against the maximal shape $s^*$. The second model, $\mathcal{M'}_{RD^{0}}$, shows good performances with an average $F1$ macro score of 0.91  and 0.97 at the micro level. 
Moreover, this model generally reproduces quite well the original patterns ($r_{\hat{\mathbb{G}}^{\leftrightarrow}_D}$) associated with the ground truth examples.
Concerning the gold model $\mathcal{M'}_{RD^{1+}}$ it is naturally more adapted to reproduce the triples of $RD^{2+}$, as it was trained on a corrected dataset: it records the best $F1$ metrics, low $\overline{Loss(\hat{\mathbb{G}})}$. 
Finally, we can notice the high rate of generated triples following the initial patterns ($r_{\hat{\mathbb{G}}^{\leftrightarrow}_D}$): twice more than $\mathcal{M}_{RD^{-}}$ and 10\% more than $\mathcal{M'}_{RD^{0}}$.

\section{Error analysis}\label{sec:error_analysis}

\subsubsection{Pattern errors analysis.} 
We consider the set $\widehat{\mathbb{G}}^{\nleftrightarrow}_{D} $ of the predicted $\hat{g}$ whose properties do not correspond to the set of expected properties $\mathcal{P}(g)$, and the corresponding ratios:
\begin{equation}\widehat{\mathbb{G}}^{\nleftrightarrow}_{D}  = \{ \hat{g} ; (w,g) \in D ; \hat{g} \not\leftrightarrow \mathcal{P}(g) \}
\qquad
 r_{\widehat{\mathbb{G}}^{\nleftrightarrow}_{D}}=\frac{|\widehat{\mathbb{G}}^{\nleftrightarrow}_{D}|}{|D|} \qquad
 r_{\widehat{\mathbb{G}}^{\nleftrightarrow}_{D}}=1-\-r_{\widehat{\mathbb{G}}^{\leftrightarrow}_{D}} 
 \end{equation}
 
We also define in the same way $\widehat{\mathbb{G}}^{\rightarrow}_{D}$, $r_{\widehat{\mathbb{G}}^{\rightarrow}_{D}}$, etc. We combine the subsets to measure the pattern extension capacity on a dataset, $PEC_D$, 
which is the ratio of predicted graphs  that strictly extend the expected pattern:
\begin{equation} 
\begin{split}PEC_D=\frac{|\widehat{\mathbb{G}}^{\rightarrow}_{D}|-|\widehat{\mathbb{G}}^{\leftrightarrow}_{D}|}{|\widehat{\mathbb{G}}^{\nleftrightarrow}_{D}|}
\end{split}
\end{equation}
\\
\noindent Finally we extend the definition of eq.~\ref{eq:pattern-set-k-s} to consider the patterns found in the graph $\hat{g}$ inferred w.r.t the ground truth $(w,g)$ in a dataset $D$:
\begin{equation} \label{eq:Patterns-D-s}
\begin{split}
\widehat{\mathbb{P}}_{D}(s) \coloneqq \{\pi \in \mathbb{P}(s); \exists (w,g) \in D ; \hat{g} \leftrightarrow \pi\}
\end{split}
\end{equation}

The resulting notation allows us to compare 
the pattern set represented in the expected graph and the predicted one:
\begin{equation} \label{eq:AZE}
\begin{split}
\mathbb{P}_{D}^{\nleftrightarrow} \coloneqq \{\pi \in \Pi; \exists (w,g) \in D ; \pi\coloneqq\mathcal{P}(g); \hat{g} {\nleftrightarrow} \pi \} \\
\widehat{\mathbb{P}}_{D}^{\nleftrightarrow} \coloneqq \{\pi \in \Pi; \exists (w,g) \in D ; \pi\coloneqq\mathcal{P}(\hat{g}); g {\nleftrightarrow} \pi \}
\end{split}
\end{equation}
For instance, $\mathbb{P}_{RD^1}(s^*)$ represents the set of patterns that can be built from $s^*$ and found in the graph set from $RD^1$, and $\widehat{\mathbb{P}}_{RD^1}(s^*)$ the set of patterns found in the predictions obtained from $RD^1$. \\

\noindent Moreover we also extend eq.~\ref{eq:r_s_D} to compute the predicted $
r_{s^*}(\hat{\mathbb{G}}^{\nleftrightarrow}_D)$ and the expected $
r_{s^*}(\mathbb{G}^{\nleftrightarrow}_D)$, that is, the rates of graphs valid against the maximal shape $s^*$, respectfully for the predictions and the expected values.

\begin{table}[h!]
\centering
\setlength{\tabcolsep}{0.6em}
\begin{tabular}{l|llllll}
\hline
$\mathcal{M}$odel (dataset D)       & $r_{\hat{\mathbb{G}}^{\nleftrightarrow}_D}$ & $|\mathbb{P}_{D}^{\nleftrightarrow}|$ & $|\widehat{\mathbb{P}}_{D}^{\nleftrightarrow}|$  &  $r_{s^*}(\mathbb{G}^{\nleftrightarrow}_D)$ & $r_{s^*}(\hat{\mathbb{G}}^{\nleftrightarrow}_D)$ & $PEC_D$  \\
\hline
$\mathcal{M}_{RD^{-}}(RD^{2+})$   & \textbf{0.57}          & \textbf{30.5} & \underline{10.2} & \underline{0.29}        & \textbf{1.00}        & 0.56 \\
\hline
$\mathcal{M'}_{RD^{0}}(RD^{2})$   & 0.27          & 28.90 & 19.10 & \textbf{0.67}        & \underline{0.53}        & \underline{0.18} \\
$\mathcal{M'}_{RD^{0}}(RD^{2+})$  & 0.28          & \underline{27.70} & \textit{23.50} &\textit{0.57}        & 0.62        & 0.58 \\
\hline
$\mathcal{M'}_{RD^{1+}}(RD^{2})$  & \textit{0.38}          & \textit{29.10} & 20.90 & 0.38        & 0.59        & \textit{0.66} \\
$\mathcal{M'}_{RD^{1+}}(RD^{2+})$  & \underline{0.20}          & 28.90 & \textbf{27.60} & 0.46        & \textit{0.73}        & \textbf{0.87}
 \end{tabular}
 \caption{Focus on the triples generated not following the initial property signature ($\hat{g} \not\leftrightarrow \mathcal{P}(g)$ )}
 \label{tab:results_3}
 \vspace{-20pt}
\end{table}

\noindent From Table~\ref{tab:results_3}, we can first notice the inability of $\mathcal{M}_{RD^{-}}$ to reproduce correctly the graphs that were not originally entailing the maximal shape $s^*$ (cf. $r_{s^*}(\mathbb{G}^{\nleftrightarrow}_D)$ ). Moreover the patterns generated by $\mathcal{M}_{RD^{-}}$  are less varied ($|\widehat{\mathbb{P}}_{RD^{-}}^{\nleftrightarrow}|$ ) than the one expected ($|\mathbb{P}_{RD^{-}}^{\nleftrightarrow}|$) but their are by design all following the maximal shape $s^*$. Conversely, the generated triples of $\mathcal{M'}_{RD^{0}}$ ($|\mathbb{P}_{RD^{0}}^{\nleftrightarrow}|$) are closer to the pattern of the expected graph ($|\widehat{\mathbb{P}}_{RD^{0}}^{\nleftrightarrow}|$ ). The correction of $RD^2$ had a slight impact on the number of triples valid against the shape, but it shows the potential in terms of pattern extension of $\mathcal{M'}_{RD^{0}}$ by reaching the same levels than $PEC_{RD^{-}}$. In addition, our gold model $\mathcal{M'}_{RD^{1+}}$  produces pattern closer to $\mathbb{P}_{RD^{1+}}^{\nleftrightarrow}$ and also tends to produce triples closer to $s^*$. Finally, this gold model obtains a high $PEC_D$, promising for knowledge completion. \\

\begin{table}[h!]
\centering
\vspace{-20pt}
\setlength{\tabcolsep}{0.6em}
\begin{tabular}{r|lll|lll}
$\mathcal{M}$odel (dataset D) & $FN_-$& $FN_+$ & $r_{omis}$ & $FP_-$& $FP_+$ & $r_{disco}$  \\
\hline
$\mathcal{M}_{RD^{-}}(RD^{2+})$ & 2 & 144  & \underline{0.005} & 198 & 95 & \underline{0.32} \\
$\mathcal{M'}_{RD^{0}}(RD^{2+})$ & 2.9 & 160 & \underline{0.006} & 27 & 28 & 0.52 \\
$\mathcal{M'}_{RD^{1+}}(RD^{2+})$ & 4.5 & 49.7 & \textbf{0.01} & 33 & 70 &\textbf{0.68} \\
\end{tabular}
\caption{Metrics computed after annotation and averaged over the 10.folds of each model}
  \label{tab:annot_analysis}
  \vspace{-40pt}
\end{table}

\subsubsection{Annotation of the FP/FN triples.} The annotator was asked to verify the false positives (FP) and false negatives (FN) triples produced by each model when tested on $RD^{2+}$.
Annotating the 10-folds of FN/FP triples took 30 to 40 minutes per model, depending on the sample size, plus a further 20 to 40 minutes to classify the mentioned errors.
We obtained four sets of triples from our annotation: the erroneous FP- (which could be linked to an error or a hallucination), the discoveries FP+, the correct False negatives FN- and the omissions FN+. We build two metrics from them: the omission rate, relative to the number of the total expected triples ($Nb_{triple\:expected}$) and the rate of discoveries, which consider the number of FP generated that could be considered as relevant, defined as follows:\\
$$r_{omis}=\frac{|FN_+|}{Nb_{triple\:expected}}  \;\;\;\;\;\; r_{disco}=\frac{|FP_+]}{|FP_+| + |FP_-|}$$

Table~\ref{tab:annot_analysis} shows us the FP produced by $\mathcal{M}_{RD^{-}}$ are mainly not relevant, by recording at the same time a high omission rate. Inversely, $\mathcal{M'}_{RD^{1+}}$ is very interesting because this one rarely omits facts, and more than half of the $FP$ produced by the models could be considered discoveries. Beyond the analysis of the ratios, if we consider the total number of FP produced, we see that model $\mathcal{M'}_{RD^{1+}}$ produces more FP+ than $\mathcal{M'}_{RD^{0}}$ however the number of erroneous one ($FP-$) remains more or less the same. 
 \vspace{-10pt}

\subsubsection{$FP-$ classification.}\label{sec:fp-classif} 
\begin{figure}[H]
\vspace{-20pt}
\begin{minipage}[t]{.5\linewidth}
  \includegraphics[width=\columnwidth]{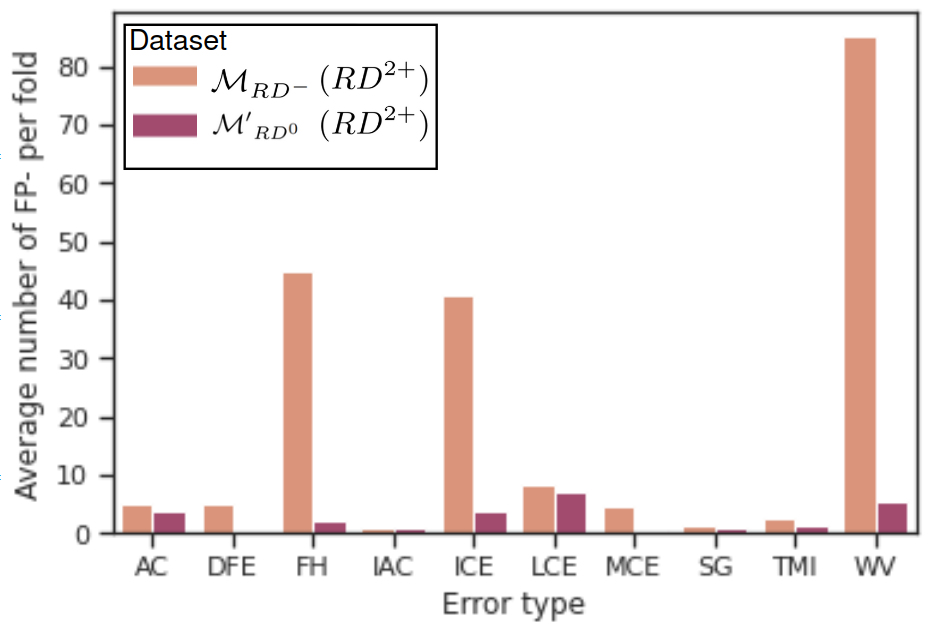}
  \label{fig:annotation_errors_M0}
\end{minipage}
\begin{minipage}[t]{.5\linewidth}
  \includegraphics[width=\columnwidth]{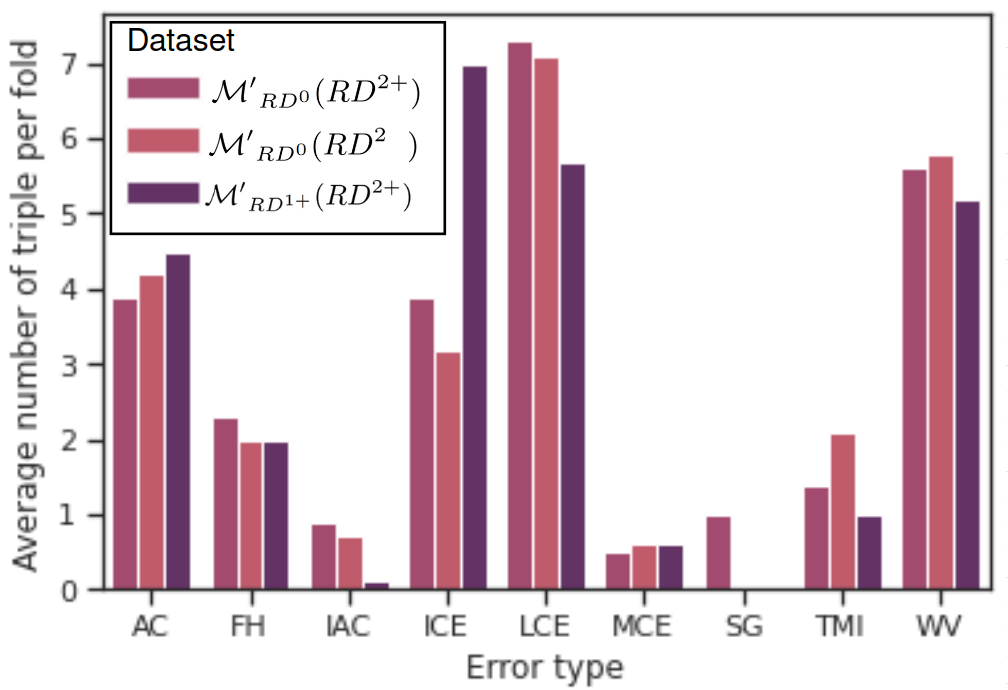}
  \label{fig:annotation_errors_others}
\end{minipage}
\vspace{-20pt}
\caption{Errors distribution over the different models and datasets}
\label{fig:error_distrib}
\vspace{-20pt}
\end{figure}

In a second step, the annotator was asked to categorize the $FP^-$ (i.e. the erroneous true positives) depending on the following classification. To illustrate it, an example of an error is given in Appendix~\ref{annex:ExampleErrors}, and the resulting distribution of these categories obtained over the testing sets $RD^2$ and $RD^{2+}$ is drawn in Figure~\ref{fig:error_distrib}:
\begin{itemize}
    \item FH - Factual hallucination: the value generated respects the range of the property, e.g. the model output is a year attached to a birthyear property, but this value is not in the abstract and could not be inferred from it.  
    \item AC - Abusive completion: the value generated respects the range of the property, and a part of this sequence is in the text, but the model completes this sequence with plausible tokens that could not be inferred by the text, e.g. the value is a date, but we could only deduce the year from the abstract).
    \item IAC - Illogical and abusive completion: The generated output is an abusive completion, and the resulting value does not respect the expected range, e.g. a date containing a day superior to 31)
    \item WV - Wrong value: The generated output is in the text but does not correspond to the targeted predicate, e.g. The value generated is related to the alias when it comes to predicting the label.
    \item TMI - Typographic minor issue: The generated output is close to the expected one, but it contains a minor typographic difference, e.g. URL encoding errors, uppercase, missing space or special characters.
    \item SG - Stuttered generation: The output is almost correct, but it contains repeated patterns, e.g. the value returned is a birthname repeated two times.
    \item ICE - Incomplete Context Error: The output is in the text and corresponds only to a part of the expected value, e.g. the models return a shortened composed birthname. 
    \item LCE - Larger context error: the output is in the text, and the expected value is inside it, but the output contains too much information. e.g. The expected label is an abbreviated birthname, and the prediction is the entire birthname.
    \item MCE - Mixed context error: the output is false but close to the expected one, and the produced value is composed of a mix of values found in the text that is not the expected one, e.g. The date return is the mix of two dates found in the text.
\end{itemize}

First of all, the bar chart on the left shows a huge gap between $\mathcal{M}_{RD^{-}}$  and $\mathcal{M'}_{RD^{0}}$. In fact, $\mathcal{M}_{RD^{-}}$  generates considerably more errors, with a majority of wrong values (WV), a lot of hallucinations (FH) and many incomplete context errors (ICE). 
The bar chart on the right shows that our new models are more likely to generate abusive completion than hallucination (AC vs. FH). However, the errors remain more or less the same despite our annotation. 

To address remaining errors, the NLI and the Triplet Critic models~\cite{huguet-cabot-navigli-2021-rebel-relation,huguet-cabot-etal-2023-red,Rogulsky2024TheEO} could be applied to each triple of $\hat{g}$. But when we applied these models to the manually evaluated FP and FN triples obtained from $\mathcal{M'}_{RD^{1+}}$ on $RD^{2+}$ we showed they do not perform well on that task (see Appendix~\ref{annex:NLI}). This can be explained by the fact that we are focusing on particular datatype properties. These observations highlight the need to adapt such models to efficiently filter potential hallucinations or recurrent errors highlighted during our analysis. In future work, we shall consider integrating the annotations into the knowledge base, which may help adapt models with contrastive learning approaches after parsing.

\section{Discussion} 

\textbf{Scalability and extensibility: } The shape complexity that can be handled by our framework is firstly limited by the output size of the SLM, which forces us to focus only on a reduced set of properties. In DBpedia, resources of type \texttt{dbo:Person} can described by up to 139 datatype properties, among which exactly 100 are used, but only 25 of them are used in more than 1\% of the cases. We chose to focus on the 7 properties that are most likely to be found in the abstracts. We observed that only half of the possible combinations (70/127) exist in our KB after Wikicheck. This remains a manageable number of combinations. Shapes containing a set of less popular properties will lead to an even smaller number of combinations. 
Additionally, considering approaches based on BART-large, such as REBEL, which can scale to up to 220 types of relations, we might expect to reach similar capabilities with a larger model than our current BART-base model. 
Considering that the shape targeting \texttt{dbo:Person} resources already cover 1/6th of the DBpedia entities, this demonstrates that Kastor can scale to datasets with large numbers of similar instances. In that context our framework is generic and could be adapted to other use cases, relying on a dual base (KG+TXT) and a SHACL shape containing only datatype properties. 

\textbf{A light active learning: }
We observed that correcting the training set with a single loop and a unique expert annotator is enough to increase the quality of the graphs produced. This annotation iteration of the errors was conducted on 10-folds, thus collecting the errors from 10 different models. This allows us to cover a lot of cases that can occur with datatype properties, although more marginal types of error may still occur. Conversely, further extension to object properties would require extending our typology of errors. We could envisage having several iterations for active learning and using pattern extension capacity or the discovery rate as a stopping criterion. In practice, we would only perform such iterations on the best model. From our point of view, one annotator and one iteration are enough as they increase the F1 performance by 10\%. However, this does not reduce the remaining marginal errors, which suggests that iterating would have little effect. Moreover, we protect our setup from noise with different strategies: the unique expert evaluator, the strict evaluation of the values and the 10-folds annotation. Adding more evaluators would introduce noise and be more costly.

\section{Conclusion}\label{sec:conclusion}

We presented Kastor, an open, reusable and extendable framework to perform an RDF-pattern relation extraction task from a noisy and incomplete knowledge base\footnote{
code : \url{https://github.com/datalogism/Kastor}\\
models: \url{https://zenodo.org/records/14498940}\\
dataset \url{https://zenodo.org/records/14382674}}.  Our approach firstly demonstrates its frugality: from the model's fine-tuning aspect that requires less than 10 minutes to a light active learning process, implying an annotator on a small set of FP/FN triples. Concerning our first research question (\textbf{RQ1}), we showed that using example-specific achievable patterns improves the performance of the relation extraction model by almost 10\% in terms of $F_1^+$. Moreover, it produces a wider variety of property patterns by avoiding many of the hallucinations plaguing the original design which relied only on graphs valid against a maximal SHACL shape.  Additionally, regarding the second research question (\textbf{RQ2}), the impact of the active learning process was also demonstrated. It leads to a better model in terms of $F_1$ scores, better ability to generate diverse patterns, better pattern extension capacities, and a better chance of generating discoveries, i.e. facts that are relevant but not initially present in the KB. Kastor is opening the door to many possible future works: the characterisation of the RDF-pattern distribution gives us the opportunity to better deal with the long-tail thereof. 
Moreover, the framework's generalizability also allows us to reproduce this current work on any SHACL shape focused on datatype properties and, with further development and system adaptation, on object properties.

\paragraph{\textbf{Acknowledgments}}
This work is supported by  
3IA Côte d'Azur (ANR-19-P3IA-0002), UCAJEDI (ANR-15-IDEX-01),  
the OPAL infrastructure and Université Côte d’Azur’s Center for High-Performance Computing.

\bibliographystyle{splncs04}
\bibliography{custom}

\begin{thebibliography}{10}
\providecommand{\url}[1]{\texttt{#1}}
\providecommand{\urlprefix}{URL }
\providecommand{\doi}[1]{https://doi.org/#1}

\bibitem{alt-etal-2020-tacred}
Alt, C., Gabryszak, A., Hennig, L.: {TACRED} revisited: A thorough evaluation of the {TACRED} relation extraction task. In: Jurafsky, D., Chai, J., Schluter, N., Tetreault, J. (eds.) Proceedings of the 58th Annual Meeting of the Association for Computational Linguistics. pp. 1558--1569. Association for Computational Linguistics, Online (Jul 2020). \doi{10.18653/v1/2020.acl-main.142}

\bibitem{10.1007/978-3-319-13704-9_3}
Augenstein, I., Maynard, D., Ciravegna, F.: Relation extraction from the web using distant supervision. In: Janowicz, K., Schlobach, S., Lambrix, P., Hyv{\"o}nen, E. (eds.) Knowledge Engineering and Knowledge Management. pp. 26--41. Springer International Publishing, Cham (2014)

\bibitem{dligach-etal-2022-exploring}
Dligach, D., Bethard, S., Miller, T., Savova, G.: Exploring text representations for generative temporal relation extraction. In: Naumann, T., Bethard, S., Roberts, K., Rumshisky, A. (eds.) Proceedings of the 4th Clinical Natural Language Processing Workshop. pp. 109--113. Association for Computational Linguistics, Seattle, WA (Jul 2022). \doi{10.18653/v1/2022.clinicalnlp-1.12}

\bibitem{Efeoglu2024RetrievalAugmentedGR}
Efeoglu, S., Paschke, A.: Retrieval-augmented generation-based relation extraction. ArXiv  \textbf{abs/2404.13397} (2024), \url{https://api.semanticscholar.org/CorpusID:269292881}

\bibitem{DBLP:conf/text2kg/GallardoCCHB24}
Gallardo, A.P., Consoli, S., Ceresa, M., Hulsman, R., Bertolini, L.: On constructing biomedical text-to-graph systems with large language models. In: Tiwari, S., Mihindukulasooriya, N., Osborne, F., Kontokostas, D., D'Souza, J., Kejriwal, M., Pellegrino, M.A., Rula, A., Gayo, J.E.L., Cochez, M., Alam, M. (eds.) Joint proceedings of the 3rd International workshop on knowledge graph generation from text {(TEXT2KG)} and Data Quality meets Machine Learning and Knowledge Graphs {(DQMLKG)} co-located with the Extended Semantic Web Conference {(} {ESWC} 2024), Hersonissos, Greece, May 26-30, 2024. {CEUR} Workshop Proceedings, vol.~3747, p.~12. CEUR-WS.org (2024), \url{https://ceur-ws.org/Vol-3747/text2kg\_paper10.pdf}

\bibitem{geng-etal-2023-grammar}
Geng, S., Josifoski, M., Peyrard, M., West, R.: Grammar-constrained decoding for structured {NLP} tasks without finetuning. In: Bouamor, H., Pino, J., Bali, K. (eds.) Proceedings of the 2023 Conference on Empirical Methods in Natural Language Processing. pp. 10932--10952. Association for Computational Linguistics, Singapore (Dec 2023). \doi{10.18653/v1/2023.emnlp-main.674}, \url{https://aclanthology.org/2023.emnlp-main.674/}

\bibitem{grangier2024needsmallspecializedlanguage}
Grangier, D., Katharopoulos, A., Ablin, P., Hannun, A.: Need a small specialized language model? plan early! (2024), \url{https://arxiv.org/abs/2402.01093}

\bibitem{hofer2023constructionknowledgegraphsstate}
Hofer, M., Obraczka, D., Saeedi, A., Köpcke, H., Rahm, E.: Construction of knowledge graphs: State and challenges (2023), \url{https://arxiv.org/abs/2302.11509}

\bibitem{huang2023surveyhallucinationlargelanguage}
Huang, L., Yu, W., Ma, W., Zhong, W., Feng, Z., Wang, H., Chen, Q., Peng, W., Feng, X., Qin, B., Liu, T.: A survey on hallucination in large language models: Principles, taxonomy, challenges, and open questions (2023), \url{https://arxiv.org/abs/2311.05232}

\bibitem{huguet-cabot-etal-2023-red}
Huguet~Cabot, P.L., Tedeschi, S., Ngonga~Ngomo, A.C., Navigli, R.: {RED}$^{\textrm{fm}}$: a filtered and multilingual relation extraction dataset. In: Rogers, A., Boyd-Graber, J., Okazaki, N. (eds.) Proceedings of the 61st Annual Meeting of the Association for Computational Linguistics (Volume 1: Long Papers). pp. 4326--4343. Association for Computational Linguistics, Toronto, Canada (Jul 2023). \doi{10.18653/v1/2023.acl-long.237}

\bibitem{huguet-cabot-navigli-2021-rebel-relation}
Huguet~Cabot, P.L., Navigli, R.: {REBEL}: Relation extraction by end-to-end language generation. In: Moens, M.F., Huang, X., Specia, L., Yih, S.W.t. (eds.) Findings of the Association for Computational Linguistics: EMNLP 2021. pp. 2370--2381. Association for Computational Linguistics, Punta Cana, Dominican Republic (Nov 2021). \doi{10.18653/v1/2021.findings-emnlp.204}

\bibitem{gahnem2024}
Hussam~Ghanem, C.C.: Fine-tuning vs. prompting: Evaluating the knowledge graph construction with llms (2024), \url{https://ceur-ws.org/Vol-3747/text2kg\_paper7.pdf}

\bibitem{Ji_2023}
Ji, Z., Lee, N., Frieske, R., Yu, T., Su, D., Xu, Y., Ishii, E., Bang, Y.J., Madotto, A., Fung, P.: Survey of hallucination in natural language generation. ACM Computing Surveys  \textbf{55}(12),  1–38 (Mar 2023). \doi{10.1145/3571730}

\bibitem{josifoski-etal-2022-genie}
Josifoski, M., De~Cao, N., Peyrard, M., Petroni, F., West, R.: {G}en{IE}: Generative information extraction. In: Carpuat, M., de~Marneffe, M.C., Meza~Ruiz, I.V. (eds.) Proceedings of the 2022 Conference of the North American Chapter of the Association for Computational Linguistics: Human Language Technologies. pp. 4626--4643. Association for Computational Linguistics, Seattle, United States (Jul 2022). \doi{10.18653/v1/2022.naacl-main.342}, \url{https://aclanthology.org/2022.naacl-main.342}

\bibitem{10.5555/3618408.3619049}
Kandpal, N., Deng, H., Roberts, A., Wallace, E., Raffel, C.: Large language models struggle to learn long-tail knowledge. In: Proceedings of the 40th International Conference on Machine Learning. ICML'23, JMLR.org (2023)

\bibitem{DBLP:conf/esws/LehmannMMORSV24}
Lehmann, J., Meloni, A., Motta, E., Osborne, F., Recupero, D.R., Salatino, A.A., Vahdati, S.: Large language models for scientific question answering: An extensive analysis of the sciqa benchmark. In: Mero{\~{n}}o{-}Pe{\~{n}}uela, A., Dimou, A., Troncy, R., Hartig, O., Acosta, M., Alam, M., Paulheim, H., Lisena, P. (eds.) The Semantic Web - 21st International Conference, {ESWC} 2024, Hersonissos, Crete, Greece, May 26-30, 2024, Proceedings, Part {I}. Lecture Notes in Computer Science, vol. 14664, pp. 199--217. Springer (2024). \doi{10.1007/978-3-031-60626-7\_11}

\bibitem{10.1007/978-3-031-60626-7_11}
Lehmann, J., Meloni, A., Motta, E., Osborne, F., Recupero, D.R., Salatino, A.A., Vahdati, S.: Large language models for scientific question answering: An extensive analysis of the sciqa benchmark. In: Mero{\~{n}}o~Pe{\~{n}}uela, A., Dimou, A., Troncy, R., Hartig, O., Acosta, M., Alam, M., Paulheim, H., Lisena, P. (eds.) The Semantic Web. pp. 199--217. Springer Nature Switzerland, Cham (2024)

\bibitem{Li2023EvaluatingCI}
Li, B., Fang, G., Yang, Y., Wang, Q., Ye, W., Zhao, W., Zhang, S.: Evaluating chatgpt's information extraction capabilities: An assessment of performance, explainability, calibration, and faithfulness. ArXiv  \textbf{abs/2304.11633} (2023), \url{https://api.semanticscholar.org/CorpusID:258297899}

\bibitem{li-etal-2022-overcoming}
Li, D., Chen, Z., Cho, E., Hao, J., Liu, X., Xing, F., Guo, C., Liu, Y.: Overcoming catastrophic forgetting during domain adaptation of seq2seq language generation. In: Carpuat, M., de~Marneffe, M.C., Meza~Ruiz, I.V. (eds.) Proceedings of the 2022 Conference of the North American Chapter of the Association for Computational Linguistics: Human Language Technologies. pp. 5441--5454. Association for Computational Linguistics, Seattle, United States (Jul 2022). \doi{10.18653/v1/2022.naacl-main.398}

\bibitem{ijcai2024p0704}
Li, G., Wang, P., Ke, W., Guo, Y., Ji, K., Shang, Z., Liu, J., Xu, Z.: Recall, retrieve and reason: Towards better in-context relation extraction. In: Larson, K. (ed.) Proceedings of the Thirty-Third International Joint Conference on Artificial Intelligence, {IJCAI-24}. pp. 6368--6376. International Joint Conferences on Artificial Intelligence Organization (8 2024). \doi{10.24963/ijcai.2024/704}, main Track

\bibitem{Li_2023}
Li, M., Shi, T., Ziems, C., Kan, M.Y., Chen, N., Liu, Z., Yang, D.: Coannotating: Uncertainty-guided work allocation between human and large language models for data annotation. In: Proceedings of the 2023 Conference on Empirical Methods in Natural Language Processing. Association for Computational Linguistics (2023). \doi{10.18653/v1/2023.emnlp-main.92}

\bibitem{liu_are_2024}
Liu, Y., Li, D., Wang, K., Xiong, Z., Shi, F., Wang, J., Li, B., Hang, B.: Are {LLMs} good at structured outputs? a benchmark for evaluating structured output capabilities in {LLMs}. Information Processing \& Management  \textbf{61}(5),  103809 (2024). \doi{https://doi.org/10.1016/j.ipm.2024.103809}, \url{https://www.sciencedirect.com/science/article/pii/S0306457324001687}

\bibitem{lu2024smalllanguagemodelssurvey}
Lu, Z., Li, X., Cai, D., Yi, R., Liu, F., Zhang, X., Lane, N.D., Xu, M.: Small language models: Survey, measurements, and insights (2024), \url{https://arxiv.org/abs/2409.15790}

\bibitem{ma-etal-2023-dreeam}
Ma, Y., Wang, A., Okazaki, N.: {DREEAM}: Guiding attention with evidence for improving document-level relation extraction. In: Vlachos, A., Augenstein, I. (eds.) Proceedings of the 17th Conference of the European Chapter of the Association for Computational Linguistics. pp. 1971--1983. Association for Computational Linguistics, Dubrovnik, Croatia (May 2023). \doi{10.18653/v1/2023.eacl-main.145}

\bibitem{vandermeer2024annotatorcentricactivelearningsubjective}
van~der Meer, M., Falk, N., Murukannaiah, P.K., Liscio, E.: Annotator-centric active learning for subjective nlp tasks (2024), \url{https://arxiv.org/abs/2404.15720}

\bibitem{paolini2021structuredpredictiontranslationaugmented}
Paolini, G., Athiwaratkun, B., Krone, J., Ma, J., Achille, A., Anubhai, R., dos Santos, C.N., Xiang, B., Soatto, S.: Structured prediction as translation between augmented natural languages (2021), \url{https://arxiv.org/abs/2101.05779}

\bibitem{10.1007/978-3-031-78952-6_8}
Ringwald, C., Gandon, F., Faron, C., Michel, F., Akl, H.A.: 12 shades of rdf: Impact of syntaxes on data extraction with language models. In: Mero{\~{n}}o~Pe{\~{n}}uela, A., Corcho, O., Groth, P., Simperl, E., Tamma, V., Nuzzolese, A.G., Poveda-Villal{\'o}n, M., Sabou, M., Presutti, V., Celino, I., Revenko, A., Raad, J., Sartini, B., Lisena, P. (eds.) The Semantic Web: ESWC 2024 Satellite Events. pp. 81--91. Springer Nature Switzerland, Cham (2025)

\bibitem{Rogulsky2024TheEO}
Rogulsky, S., Popovic, N., Färber, M.: The effects of hallucinations in synthetic training data for relation extraction. arxiv  (2024)

\bibitem{rossiello2022knowglknowledgegenerationlinking}
Rossiello, G., Chowdhury, M.F.M., Mihindukulasooriya, N., Cornec, O., Gliozzo, A.M.: Knowgl: Knowledge generation and linking from text (2022), \url{https://arxiv.org/abs/2210.13952}

\bibitem{SHENOY2022100679}
Shenoy, K., Ilievski, F., Garijo, D., Schwabe, D., Szekely, P.: A study of the quality of wikidata. Journal of Web Semantics  \textbf{72},  100679 (2022). \doi{https://doi.org/10.1016/j.websem.2021.100679}

\bibitem{10.1145/3241741}
Smirnova, A., Cudr\'{e}-Mauroux, P.: Relation extraction using distant supervision: A survey. ACM Comput. Surv.  \textbf{51}(5) (nov 2018). \doi{10.1145/3241741}, \url{https://doi.org/10.1145/3241741}

\bibitem{Stoica2021ReTACREDAS}
Stoica, G., Platanios, E.A., P'oczos, B.: Re-tacred: Addressing shortcomings of the tacred dataset. In: AAAI Conference on Artificial Intelligence (2021), \url{https://api.semanticscholar.org/CorpusID:233296843}

\bibitem{taille-etal-2020-lets}
Taill{\'e}, B., Guigue, V., Scoutheeten, G., Gallinari, P.: Let{'}s {S}top {I}ncorrect {C}omparisons in {E}nd-to-end {R}elation {E}xtraction! In: Webber, B., Cohn, T., He, Y., Liu, Y. (eds.) Proceedings of the 2020 Conference on Empirical Methods in Natural Language Processing (EMNLP). pp. 3689--3701. Association for Computational Linguistics, Online (Nov 2020). \doi{10.18653/v1/2020.emnlp-main.301}

\bibitem{tsaneva_enhancing_nodate}
Tsaneva, S., Sabou, M.: Enhancing human-in-the-loop ontology curation results through task design. {ACM} J. Data Inf. Qual.  \textbf{16}(1),  4:1--4:25 (2024). \doi{10.1145/3626960}, \url{https://doi.org/10.1145/3626960}

\bibitem{wadhwa-etal-2023-revisiting}
Wadhwa, S., Amir, S., Wallace, B.: Revisiting relation extraction in the era of large language models. In: Rogers, A., Boyd-Graber, J., Okazaki, N. (eds.) Proceedings of the 61st Annual Meeting of the Association for Computational Linguistics (Volume 1: Long Papers). pp. 15566--15589. Association for Computational Linguistics, Toronto, Canada (Jul 2023). \doi{10.18653/v1/2023.acl-long.868}

\bibitem{wang2024comprehensivesurveysmalllanguage}
Wang, F., Zhang, Z., Zhang, X., Wu, Z., Mo, T., Lu, Q., Wang, W., Li, R., Xu, J., Tang, X., He, Q., Ma, Y., Huang, M., Wang, S.: A comprehensive survey of small language models in the era of large language models: Techniques, enhancements, applications, collaboration with llms, and trustworthiness (2024), \url{https://arxiv.org/abs/2411.03350}

\bibitem{yao-etal-2021-codred}
Yao, Y., Du, J., Lin, Y., Li, P., Liu, Z., Zhou, J., Sun, M.: {C}od{RED}: A cross-document relation extraction dataset for acquiring knowledge in the wild. In: Moens, M.F., Huang, X., Specia, L., Yih, S.W.t. (eds.) Proceedings of the 2021 Conference on Empirical Methods in Natural Language Processing. pp. 4452--4472. Association for Computational Linguistics, Online and Punta Cana, Dominican Republic (Nov 2021). \doi{10.18653/v1/2021.emnlp-main.366}

\bibitem{yao-etal-2019-docred}
Yao, Y., Ye, D., Li, P., Han, X., Lin, Y., Liu, Z., Liu, Z., Huang, L., Zhou, J., Sun, M.: {D}oc{RED}: A large-scale document-level relation extraction dataset. In: Korhonen, A., Traum, D., M{\`a}rquez, L. (eds.) Proceedings of the 57th Annual Meeting of the Association for Computational Linguistics. pp. 764--777. Association for Computational Linguistics, Florence, Italy (Jul 2019). \doi{10.18653/v1/P19-1074}

\bibitem{zaratiana-etal-2024-gliner}
Zaratiana, U., Tomeh, N., Holat, P., Charnois, T.: {GL}i{NER}: Generalist model for named entity recognition using bidirectional transformer. In: Duh, K., Gomez, H., Bethard, S. (eds.) Proceedings of the 2024 Conference of the North American Chapter of the Association for Computational Linguistics: Human Language Technologies (Volume 1: Long Papers). pp. 5364--5376. Association for Computational Linguistics, Mexico City, Mexico (Jun 2024). \doi{10.18653/v1/2024.naacl-long.300}

\bibitem{zhang2023usinglargelanguagemodels}
Zhang, B., Reklos, I., Jain, N., Peñuela, A.M., Simperl, E.: Using large language models for knowledge engineering (llmke): A case study on wikidata (2023), \url{https://arxiv.org/abs/2309.08491}

\bibitem{zheng2023judgingllmasajudgemtbenchchatbot}
Zheng, L., Chiang, W.L., Sheng, Y., Zhuang, S., Wu, Z., Zhuang, Y., Lin, Z., Li, Z., Li, D., Xing, E.P., Zhang, H., Gonzalez, J.E., Stoica, I.: Judging llm-as-a-judge with mt-bench and chatbot arena (2023), \url{https://arxiv.org/abs/2306.05685}

\end{thebibliography}
\newpage
\section{Appendix}

\subsection{SHACL shape}\label{annex:SHACL_SHAPE}
\begin{tiny}
\centering
\begin{multicols}{2}
\begin{verbatim}
@prefix rdf: <http://www.w3.org/1999/02/22-rdf-syntax-ns#> .
@prefix rdfs: <http://www.w3.org/2000/01/rdf-schema#> .
@prefix schema: <http://schema.org/> .
@prefix sh: <http://www.w3.org/ns/shacl#> .
@prefix xsd: <http://www.w3.org/2001/XMLSchema#> .
@prefix dbo: <http://dbpedia.org/ontology/> .


schema:PersonShape a sh:NodeShape ;
 sh:targetClass dbo:Person ; 
 sh:property [ 
   sh:path rdfs:label;
   sh:minCount 1 ;
   sh:datatype xsd:string;
 ];
sh:or (
    [
      sh:property [ 
         sh:path dbo:birthDate;
         sh:datatype xsd:date;
         sh:minCount 1;
         sh:maxCount 1;
      ]
    ]   
    [
      sh:property [ 
          sh:path dbo:birthYear;
          sh:datatype xsd:gYear;
          sh:minCount 1;
          sh:maxCount 1;
      ]
    ]
);
sh:property [ 
   sh:path dbo:deathYear;
   sh:minCount 0;
   sh:maxCount 1;
   sh:datatype xsd:gYear;
 ];
 sh:property [ 
   sh:path dbo:alias;
   sh:datatype xsd:string ;
   sh:minCount 0;
   sh:maxCount 10;
   sh:nodeKind sh:Literal;
 ];
 sh:property [ 
   sh:path dbo:birthName ;
   sh:datatype xsd:string ;
   sh:minCount 0;
   sh:maxCount 1;
   sh:nodeKind sh:Literal ;
 ] ;
 sh:property [ 
   sh:path dbo:deathDate ;
   sh:datatype xsd:date ;
   sh:minCount 0;
   sh:maxCount 1;
 ].
\end{verbatim}
\end{multicols}
\end{tiny}

\subsection{Ebnf grammar}\label{annex:EBNF}
\begin{tiny}
\begin{verbatim}
############## turtle light oneline factorized
root ::= triples+ 
triples ::= WS? triple WS? "." 
triple ::= subj WS? predicateObjectList 
predicateObjectList ::= pred objectList ( WS?  ";"  WS? ( pred   WS? objectList)? )*
objectList ::= obj  (  WS? ","  WS? obj  )* 
subj ::=  iri
pred ::= iri | "a"
obj ::= iri | string
string ::= WS? "\"" [ \t!#-\[\]-~]* "\"" WS?
iri ::= ":" PN_LOCAL+  
WS ::= [ \t\n]
PN_CHARS_BASE ::= [A-Z] | [a-z] | [#x00C0-#x00D6] | [#x00D8-#x00F6] | [#x00F8-#x02FF] |
[#x0370-#x037D] | [#x037F-#x1FFF] | [#x200C-#x200D] | [#x2070-#x218F] | [#x2C00-#x2FEF] |
[#x3001-#xD7FF] | [#xF900-#xFDCF] | [#xFDF0-#xFFFD] | [#x10000-#xEFFFF] 
PN_CHARS_U ::= PN_CHARS_BASE | "_" 
PN_LOCAL ::= ( PN_CHARS_U | ":" | [0-9] | PLX ) 
( ( PN_CHARS | "." | ":" | PLX )* 
( PN_CHARS | ":" | PLX ) ) ?
PLX  ::= PERCENT | PN_LOCAL_ESC
PN_CHARS  ::= PN_CHARS_U | "-" | [0-9] | [#x00B7] | [#x0300-#x036F] | [#x203F-#x2040] 
PERCENT  ::= "%" HEX HEX
HEX ::= [0-9] | [A-F] | [a-f]
PN_LOCAL_ESC ::= "\\" ( "_" | "~" | "." | "-" | "!" | "$" | "&" | 
"'" | "(" | ")" | "*" | "+" | "," | ";" | "=" | "/" | "?" | "#" | "@" | "%" )
\end{verbatim}
\end{tiny}
\newpage
\subsection{Example of FP- errors by category}\label{annex:ExampleErrors}
\begin{table}[]

\resizebox{\linewidth}{!}{
\begin{CJK}{UTF8}{min}
\begin{tabular}{|m{0.05\textwidth}||m{0.5\textwidth}|l|l|l|}
\hline
    & \begin{center}Context\end{center} & Predicate & Generated value & Expected value \\
    \hline
FH  & Marguerite Kathryn Flecknoe is an American voice actress, radio personality, television host and producer.                                                               & birthYear                     & \textcolor{red}{1944}                                & $\emptyset$                                   \\
\hline
AC  & Peter Woon (1931 – May 2014) was a news and current affairs editor at the British Broadcasting Corporation...                                                            & deathDate                     & 2014-05\textcolor{red}{-14}                          &                                    \\
\hline
IAC & Frederick Jardine (born 27 September 1941 died 7 october 2019 was a Scottish former professional footballer, ...                                                         & deathDate                     & 2019-\textcolor{red}{ october 2019}                  & 2019-10-07                         \\
\hline
TMI & Françoise Abanda (born February 5, 1997) is a Canadian professional tennis player. She reached her highest WTA...                                                        & label                         & Fran\textcolor{red}{\%C3\%A7oise} Abanda               & Françoise Abanda                   \\
\hline
SG  & Mao Ichimichi (市道 真央, Ichimichi Mao, born February 1, 1992) is a Japanese actress and voice actress. She started her career as a Japanese idol ...                       & birthName                     & Mao Ichimichi \textcolor{red}{Mao}                   & Mao Ichimichi                  \\
\hline
WV  & Jeremy Larroux (born 1993), better known as Laylow is a French rapper from Toulouse. In 2018, Laylow released the EPs.RAW and RAW-Z. ..                                  & alias                         & \textcolor{red}{Jeremy Larroux}                     & Laylow                             \\
\hline
ICE & Mariano Garchitorena y Chereau (February 12, 1898 - October 1, 1961) was a Filipino politician of Spanish-French descent...                                              & birthName                     & Mariano Garchitorena                & Mariano Garchitorena y Chereau     \\
\hline
LCE & Lenilson Batista de Jesús (born May 1, 1981 in Salvador), also known as Lenilson Batista de Souza, Lenilson Batista, or simply Lenílson, is a Brazilian left midfielder. & label                         & Lenilson Batista \textcolor{red}{de Jesús}           & Lenílson Batista                   \\
\hline
MCE & Stephen Edward Smith (September 24, 1927 – August 19, 1990) was the husband of Jean Ann Kennedy...                                                                       & birthName                     & \textcolor{green}{Stephen} \textcolor{red}{Ann Kennedy}                 & Stephen Edward Smith\\ 
 \hline
\end{tabular}
\end{CJK}
}
\\
\label{tab:examples_errors}
\end{table}
\newpage
\subsection{Active learning process}\label{annex:ActiveLearn}
\begin{algorithm}[tbh!]
\scriptsize
\caption{Light active learning process}\label{alg:gral}
\KwData{$\{RD^{0}, RD^{1}, RD^{2}\} \in \mathcal{K}^{\mathcal{WR}\models}_{\mathbb{P}(s^*)} $}
\everypar={\nl}
with  $RD^0 = RD^0_{test}  \sqcup  RD^0_{train}  \sqcup  RD^0_{eval}$\\
Train $\mathcal{M'}_{RD^{0}}$ using $RD^0_{train}$\;
\ForEach{ $D_i \in \{RD^{1}, RD^{2}\}$}
    {
    $FP_{D_i}  \gets \emptyset$\;
    $FN_{D_i}  \gets \emptyset$\\
    \ForEach{ $(w,g)  \in  D_i $   }{
        $\hat{g} \gets \mathcal{M}_{RD^{0}}(w)$;\\
        \ForEach{triple $\hat{t} \in \hat{g}$}{
             \If{ $\hat{t} \not\subset g$}{
                $FP_{D_i}  \gets FP_{D_i} \cup \hat{t} $
             }
        }
        
        \ForEach{triple $t \in g$}{
        \If{ $t \not\subset \hat{g}$}{
                $FN_{D_i} \gets FN_{D_i} \cup t $
        }
        }
        Human annotation of $FN_{D_i}$ and $FP_{D_i}$ as:\\
        $FN_{D_i} \gets FN^{+}_{D_i} \cup FN^{-}_{D_i}$\\
        $FP_{D_i}   \gets FP^{+}_{D_i} \cup  FP^{-}_{D_i}$\\
        Gather into $\kappa^+_i$  only valid triples:\\
        $D^+_i   \gets  (D_i  \backslash FN^{-}_{D_i} ) \cup FP^+_{D_i}$
    }
    with $RD^{1+}=D^+_1$ and $RD^{2+}=D^+_2$\\
    and  $RD^{1+} = RD^{1+}_{test}  \sqcup  RD^{1+}_{train}  \sqcup  RD^{1+}_{eval}$\\
    Train  $\mathcal{M'}_{RD^{1+}}$ using $RD^{1+}_{train}$\;
    Evaluate $\mathcal{M'}_{RD^{0}}$ and $\mathcal{M'}_{RD^{1+}}$ using $RD^{2+}$\;
 }
\end{algorithm}
\subsection{Applied NLI to annotated triples}\label{annex:NLI}
\vspace{-20pt}
\begin{figure}[h!]
    \centering
  \includegraphics[width=0.45\linewidth]{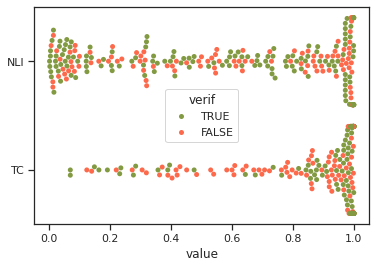}
 \caption{ $\mathcal{M'}_{RD^{1+}}(RD^{2+})$  NLI scores}
 \label{fig:NLI}
\end{figure}
\end{document}